\documentclass{article}
\usepackage[preprint]{NIPS}

\usepackage[utf8]{inputenc} % allow utf-8 input
\usepackage[T1]{fontenc}    % use 8-bit T1 fonts
\usepackage{url}            % simple URL typesetting
\usepackage{booktabs}       % professional-quality tables
\usepackage{amsfonts}       % blackboard math symbols
\usepackage{nicefrac}       % compact symbols for 1/2, etc.
\usepackage{microtype}      % microtypography

\usepackage{graphicx}
\usepackage{multirow}
\usepackage{subfigure}
\usepackage{amsmath}
\usepackage{hyperref}
\hypersetup{
    colorlinks=true,
    linkcolor=black
}

\usepackage[misc]{ifsym}

\title{Spatio-Temporal Context Modeling for Road Obstacle Detection}

\author{%
  Xiuen Wu \\
  Fuzhou University\\
  \texttt{xiuen\_wu@163.com} \\
  \And
  Tao Wang\thanks{Corresponding authors. Paper accepted by the 4th International Conference on Machine Learning for Cyber Security (ML4CS 2022), Guangzhou, China. The final authenticated version is
available online at \url{https://doi.org/10.1007/978-3-031-20096-0_17}.} \\
  Minjiang University\\
  \texttt{twang@mju.edu.cn} \\
  \And
  Lingyu Liang\footnotemark[1] \\
  South China University of Technology\\
  \texttt{eelyliang@scut.edu.cn} \\
  \And
  Zuoyong Li \\
  Minjiang University\\
  \texttt{fzulzytdq@126.com} \\
  \And
  Fum Yew Ching \\
  Universiti Sains Malaysia\\
  \texttt{fyching@student.usm.my} \\  
}

\begin{document}

\maketitle              % typeset the title of the contribution
% \index{Ekeland, Ivar} % entries for the author index
% \index{Temam, Roger}  % of the whole volume
% \index{Dean, Jeffrey}

\begin{abstract}
Road obstacle detection is an important problem for vehicle driving safety. In this paper, we aim to obtain robust road obstacle detection based on spatio-temporal context modeling. Firstly, a data-driven spatial context model of the driving scene is constructed with the layouts of the training data. Then, obstacles in the input image are detected via the state-of-the-art object detection algorithms, and the results are combined with the generated scene layout. In addition, to further improve the performance and robustness, temporal information in the image sequence is taken into consideration, and the optical flow is obtained in the vicinity of the detected objects to track the obstacles across neighboring frames. Qualitative and quantitative experiments were conducted on the Small Obstacle Detection (SOD) dataset and the Lost and Found dataset. The results indicate that our method with spatio-temporal context modeling is superior to existing methods for road obstacle detection.

\end{abstract}
\section{Introduction}
The performance of object detection algorithms has been significantly improved in recent years, largely due to the success of models based on deep convolutional neural networks~\cite{2016You,2017YOLO9000,jocher2021ultralytics,girshick2014rich,ren2015faster,cai2018cascade}.
In addition, object detection algorithms have been widely used in a spectrum of practical application scenarios. Among them, obstacle object detection in road scenes is a crucial ability for self-driving vehicles. If we can detect the obstacle in front of the vehicle ahead of time and maneuver around it, traffic collisions and consequent injuries, death, and property damage can be avoided. \par
Despite its significance in driving safety, road obstacle detection is much more challenging than generic object detection. Firstly, road obstacles can be very small especially when viewed from distance. This poses challenges to modern deep learning-based object detectors as they usually operate at fixed levels of the spatial pyramid. While we could potentially train specialized detectors for small-scale object detection, these algorithms would require additional computational budget due to the increased feature resolution. In addition, visual cues from road obstacles can be weak or ambiguous due to motion blur, illumination variation, occlusion, etc. This is particularly the case when the objects are either small or distant, or both. As humans, we overcome these difficulties by constantly focusing on road conditions while driving. An experienced driver will even know what kind of obstacles are more likely to appear at which part of the scene, and pay extra attention to these regions accordingly. Therefore, it would be ideal if an object detection method could reason about the presence of obstacles in a similar fashion. \par
\begin{figure}
\includegraphics[width=\textwidth]{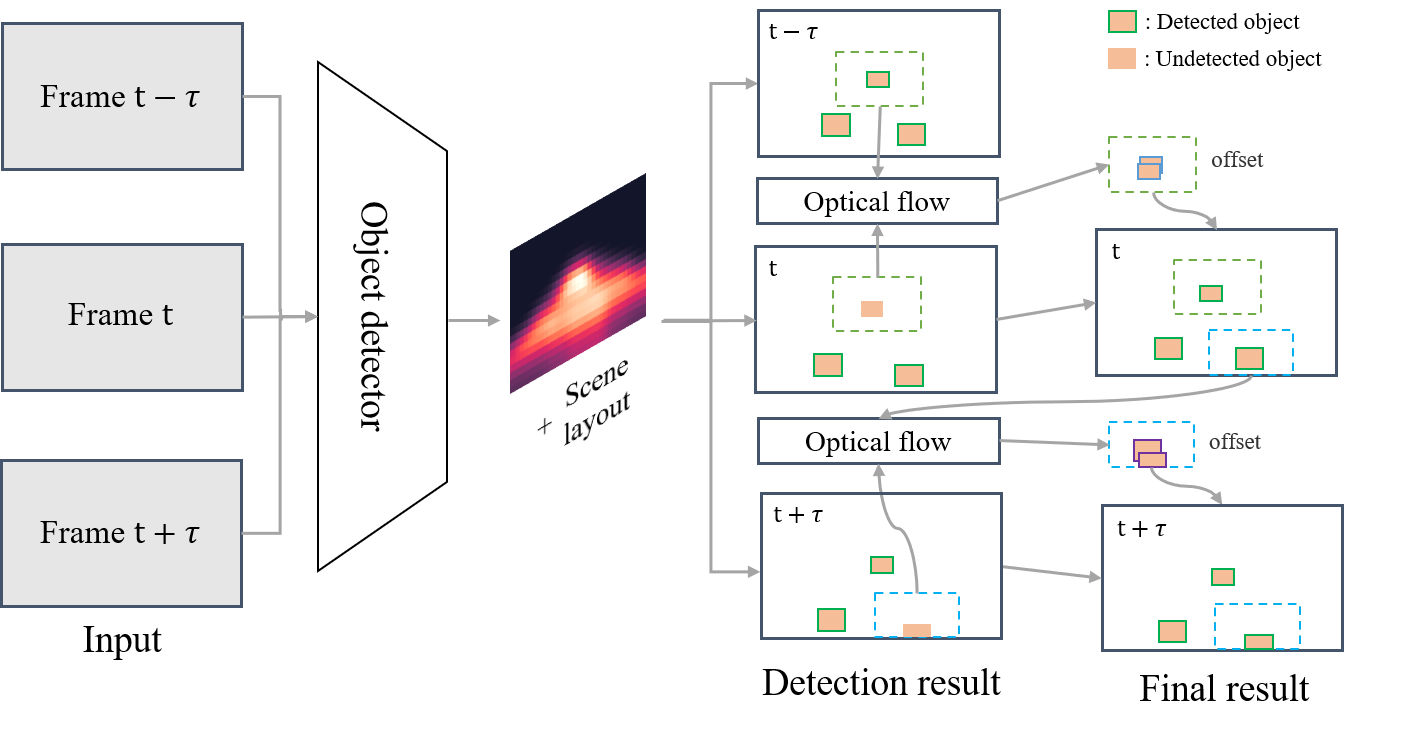}
\caption{Overview of our method. Given an input sequence of images, our method first integrates the spatial context by building a scene layout model that reasons about the spatial distribution of obstacles and road region. We then calculate the optical flow of the detected object region in two adjacent frames, so as to obtain the object position offset between them. The spatial and temporal contexts are both integrated into the final detection results.} \label{fig1}
\end{figure}

In this work, we aim to improve state-of-the-art road obstacle detectors with spatio-temporal context. For example, by analyzing the spatial context, we can discover that the obstacles are usually located on the road surface and mainly in front of the vehicle. Therefore, we adopt a data-driven approach to extract the location of obstacles and the road layout from training images to build a scene layout model, so that our detector can focus more on the objects on the road surface and eliminate the irrelevant false positives from the background region. In addition, temporal cues are also important for road obstacle detection. Sometimes, an object is detectable in one frame, but will become undetectable in the next frame due to adverse factors such as motion blur, illumination changes, or partial occlusion. Therefore, we could exploit the temporal information in consecutive frames to further assist object detection. Specifically, we use optical flow~\cite{bouguet2001pyramidal} to transfer object cues from one frame to another. We note that, in our case, the obstacles and the vehicle are both moving while driving. Therefore, when computing the optical flow between the preceding and the following frames, the flow may be too noisy. Therefore, we propose a method based on object region selection, which only calculates the optical flow of the region around the detected objects. This not only removes the complex background, but also reduces the computational complexity. See Fig.~\ref{fig1} for a high-level overview of our method. The contribution of this work is four-fold:
\begin{itemize}
\item We propose to integrate spatial and temporal context into state-of-the-art object detection algorithms based on deep learning to improve detection accuracy for road obstacles.

\item Based on the spatial context of the obstacles and roads, we construct an interpretable and effective data-driven road scene layout model.

\item Based on the temporal context, we track the detected objects via optical flow to assist their detection in the subsequent frames. In particular, we propose an optical flow calculation method based on object region selection.

\item We empirically demonstrate the superiority of our method over state-of-the-art object detection methods on the Small Obstacle Detection (SOD) dataset~\cite{singh2020lidar} and the Lost and Found dataset~\cite{pinggera2016lost}.
\end{itemize}

\section{Related Work}
\subsection{Object Detection} 
State-of-the-art methods for object detection are mainly based on Convolutional Neural Networks~(CNNs). On a broad level, these algorithms can be categorized into two-stage detectors and one-stage detectors. The two-stage detectors mainly follow the R-CNN~\cite{girshick2014rich} pipeline that includes the object proposal stage and the detection stage. The original R-CNN uses CNNs for object classification on object proposals generated by selective search~\cite{uijlings2013selective}. However, it is too slow for real-time applications due to the repeated computations of convolutional features for each proposal. Fast R-CNN~\cite{girshick2015fast} is faster than R-CNN because it performs feature extraction for object classification only once for all the region proposals. Furthermore, Faster R-CNN~\cite{ren2015faster} combines object proposal and classification into an unified model, so as to improve the efficiency of the network and allow end-to-end training. On the other hand, the YOLO series~\cite{2016You,2017YOLO9000,jocher2021ultralytics} is a classic example of one-stage detectors, which frames object detection as a regression problem. 
YOLO can directly predict the bounding boxes and the object categories without proposal generation and region refinement, so it is better suited for real-time applications. In this work, we choose Faster RCNN~\cite{ren2015faster} and YOLOv5~\cite{jocher2021ultralytics} as our baseline object detectors and explore how to improve their results with spatio-temporal context.

\subsection{Spatial Context} 
Modeling the spatial context for object detection is a well-studied problem in computer vision. For instance, Zhang et al.~\cite{zhang2014fast} construct the temporal and spatial relationship between the object and the surrounding context through the Bayesian framework. Yao et al.~\cite{yao2012describing} propose a holistic scene understanding model that simultaneously solve the problems of object detection, segmentation and scene classification. Wang et al.~\cite{wang2017efficient} propose an efficient scene layout aware object detection method for traffic surveillance. Unlike existing work, we propose a simple but effective method to construct a scene layout model that is tailored to the task of road obstacle detection. In particular, our method considers the spatial distribution of both obstacles and road region, which is not adequately investigated in the literature. 

\subsection{Temporal Context} 
There have also been papers on how to exploit the temporal context for video object detection. For example, Kang et al.~\cite{kang2017t} propose a deep learning framework to solve the problem of general object detection in videos by combining temporal and contextual information. Zhu et al.~\cite{zhu2017deep} utilize optical flow to propagate feature across frames to avoid costly feature extraction for non-key frames. Galteri et al.~\cite{galteri2017spatio} propose a closed-loop framework that uses the object detection results on the previous frame to feed back to the proposal algorithm to improve detection accuracy. Again, our approach differs from these methods as we specifically consider the challenging scenario of road obstacle detection in autonomous driving, and that we propose a method based on object region selection that limits the adverse impact of background noise.

\subsection{Road Obstacle Detection} 
In recent years, a lot of work has been done on road obstacle detection. Kyutoku et al.~\cite{kyutoku2011road} propose a method based on image subtraction, which mainly uses the difference between the road surface region of the present and past in-vehicle camera images to detect obstacles. Levi et al.~\cite{levi2015stixelnet} propose to treat the obstacle detection problem as a column-wise regression problem, and then use CNN to solve it. Leng et al.~\cite{leng2019robust} present a method that utilizes the U-V disparity map and contextual information to detect obstacles. Our work differs from the methods above in the sense that we integrate a cross-image spatial context model of the obstacles and the road into object detection, in addition to the temporal cues between consecutive frames to more accurately detect objects.

\section{Method}
In this work, we propose a general framework for integrating the spatio-temporal context into object detection, and it works with any object detection algorithm that outputs bounding boxes. Firstly, the object detection results are obtained through the detector, and then a spatial context score is calculated for the position of each bounding box using the scene layout model. This score is combined with the detection to suppress the false positives in the background. Finally, the object is tracked through the optical flow, so as to provide additional support for object hypotheses in frames with weaker visual cues. The overall process is shown in Fig.~\ref{fig1}.\par
More formally, suppose at time $t$ we have an image $I_t$ as input. Let the object hypothesis be ${x\in X}$, where $X$ is the object pose space. To simplify the notation, we assume each hypothesis is $x =(x_c,b_s,b_r,o) $ where $x_c = (b_x,b_y)$ is the image coordinate location of the object center, $b_s = (b_w, b_h) $ a scale, $b_r$ an aspect ratio and ${o\in O}$ a target class. Note that each $x$ now implies a bounding box as well. Object detection algorithms define a scoring function $S_D(x)$ for each valid object hypothesis $x$. For example, in Faster RCNN, this score is usually obtained via a multi-class softmax score on a convolutional feature map ${
f_{CNN}}$,i.e., $S_D(x) = \frac{exp(f_{CNN}(x))}{\sum\nolimits_{o\in O}exp(f_{CNN}(x))}$. We propose an additional scene layout score $S_L(x)$ for any given object hypothesis $x$. The final detection score is a weighted sum of the two scores:
\begin{equation}
 S(x) = S_D(x) + \theta S_L(x) 
\end{equation} 
where ${\theta}$ is a hyperparameter for the relative importance between the two terms. 
\par

In addition, as the image of the current frame is $I_t$, we define the image of the next frame as ${I_{(t+ \delta)}}$. Likewise, for a given object hypothesis $x$ in the current frame, the bounding box generated by optical flow tracking in the next frame is denoted as ${ x_{(t+ \delta)} }$. For the bounding box generated by optical flow, we define its scoring function in Eqn.~\ref{finalSorceByGen}.
%\begin{equation}
%S(x_{(t+ \delta)}) = S(x) + S_F(x_{(t+ \delta)})
%\end{equation}   
%\par
%In addition, we define the image of the preceding frame as $F_t$ and the image of the next frame as ${F_{(t+ \delta)}}$.The bounding box generated by optical flow tracking in the next frame is ${ x^ \prime }$.

\subsection{Scene Layout Construction} 
We obtain the spatial context information in the road scene by constructing the scene layout, and then use the obtained contextual information to provide regularization to the detection results. For the locations with high probability of obstacles, a higher detection score is given, so as to enhance the confidence of the objects that are difficult to detect on the road. In addition, by modeling the spatial distribution of roads, we can quickly eliminate the false positives outside the road region. Therefore, we build the scene layout based on the spatial distribution of both the obstacles and the road, as outlined below.

\begin{figure}
\includegraphics[width=\textwidth]{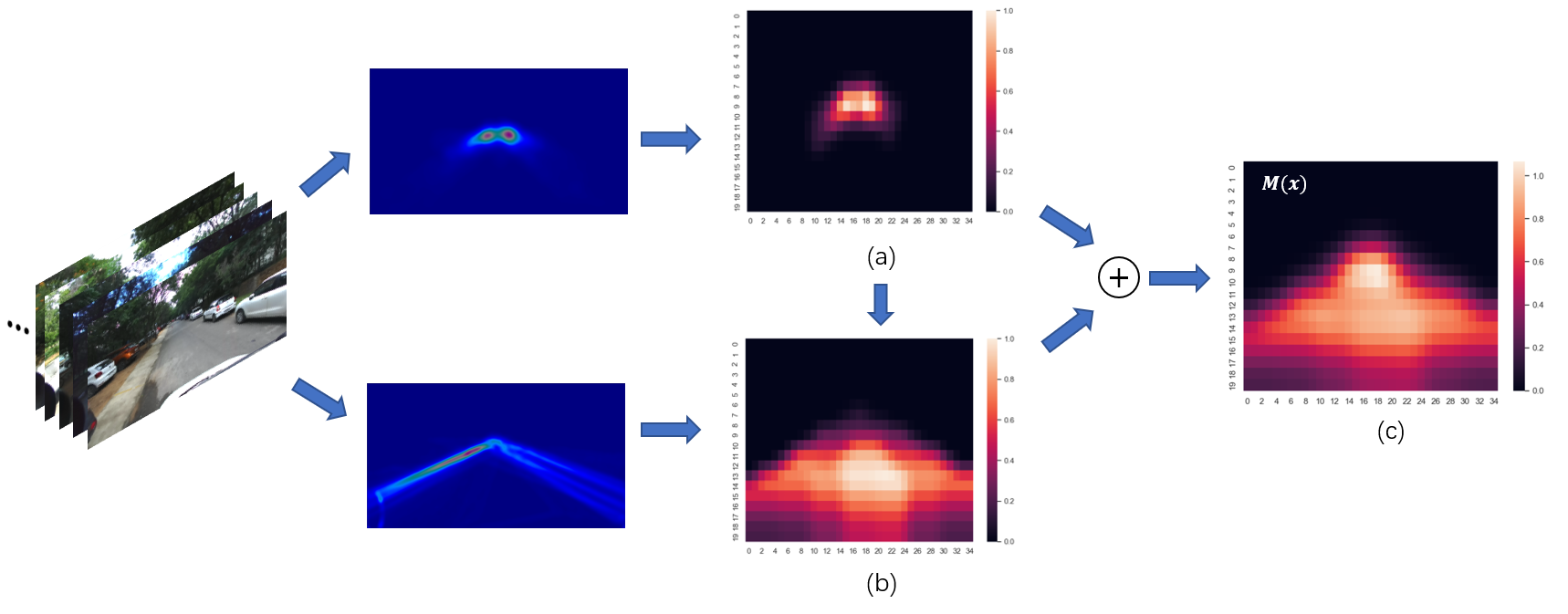}
\caption{Scene layout construction. We derive data-driven 2D distributions of obstacles and roads from the training set. (a) Obstacle distribution heat map. (b) Road distribution heat map. (c) The scene layout obtained by combining obstacle distribution and road distribution.} \label{fig2}
\end{figure}

\subsubsection{Obstacle Distribution} 
In autonomous driving, the spatial distribution of obstacles shows strong regularity, because we will usually focus on the objects in front of the vehicle as they potentially affect our driving safety. Due to the distance of these objects, they may be small and their visual cues may be weak due to motion blur, illumination, or partial occlusion. Therefore, the detector often gives low confidence for these objects and they cannot be reliably detected in practice. %\par
To address the above problem, we derive statistics on the distribution of obstacles in the training set. We first obtain all the ground truth bounding boxes in the training set and take all their center points to obtain a 2D spatial distribution in the form of a heat map. The results are shown in Fig. 2 (a). The distribution of obstacles in the image shows a strong regularity, which is mainly concentrated in the center of the image. Therefore, we use this distribution as prior information to regularize the object detection results.\par
\begin{figure}
\includegraphics[width=\textwidth]{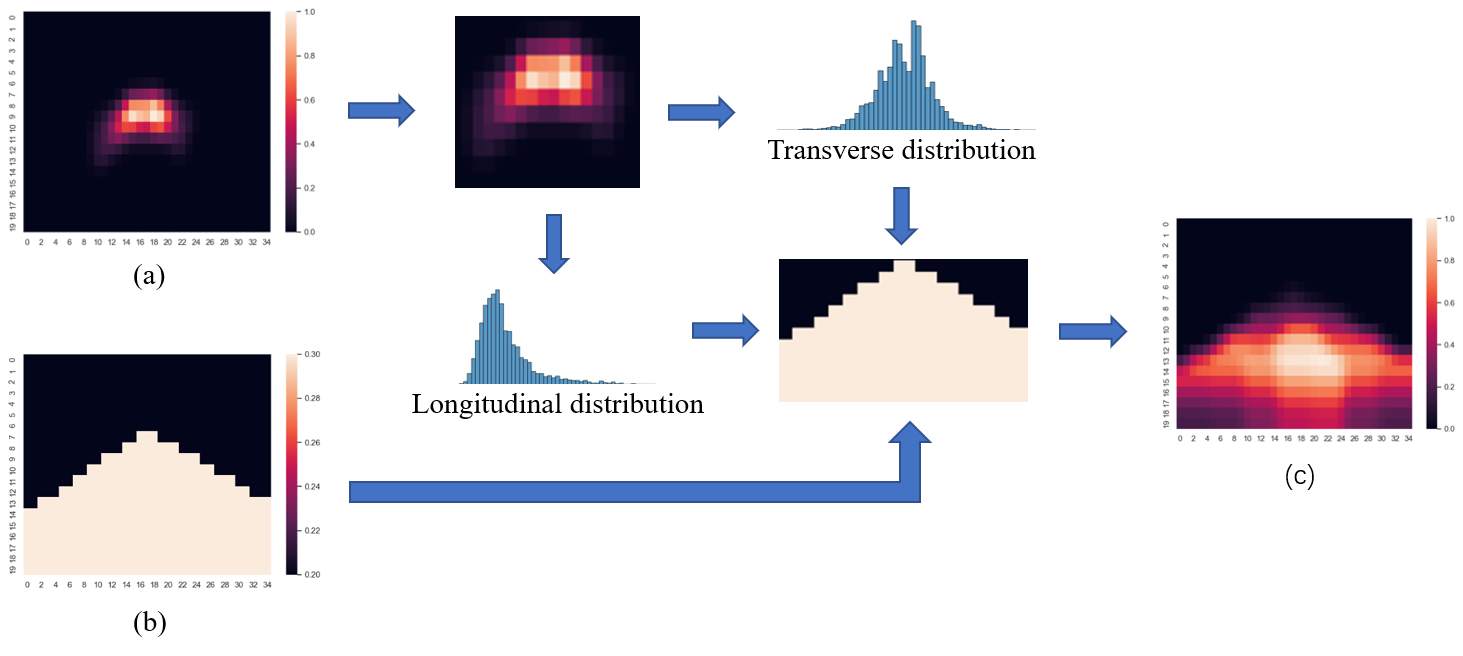}
\caption{An example of road distribution. (a) Obstacle distribution heat map. (b) Road region. (c) Road distribution obtained by spreading out the obstacle distribution along transverse and longitudinal directions in the road region.} \label{fig3}
\end{figure}
\subsubsection{Road Distribution} 
In addition, the road distribution is obtained to further assist the construction of the scene layout model, so that the obstacles in the road can be effectively detected, and the false positives outside the road region can be appropriately ignored. We note that both the SOD dataset and the Lost and Found dataset provide road segmentation annotations, so we could easily obtain the road contour in the training set. To obtain the final road contour, we take several points along the road contour in each image to obtain an average road contour across multiple images.
Since the obstacle distribution we obtained in the previous step can be unevenly distributed in the road region, here we propose a method to spread out the obstacle distribution to the entire road region. Specifically, according to the distribution of obstacles, different weights are given along the transverse and longitudinal directions in the road region, respectively. The process is illustrated in Fig.~\ref{fig3}. It can be seen that for our daily driving scenarios, the road gradually opens up from far to near, and is roughly in the shape of a pyramid. Also, we note that it is possible to design more complex road models for diverse driving scenarios, but the current model is sufficient for the relatively simple road distribution in the datasets we use in this paper.\par

\subsubsection{Scene Layout Aware Detection} 
The obstacle distribution and road distribution are added up and then normalized to obtain the final scene layout, as shown in Fig. 2 (c). Afterwards, we can generate a score $S_L(x)$ for each object hypothesis $x$ using the scene layout. Then we can combine the score of object hypothesis $x$ obtained from the object detector $S_D(x)$  with the scene layout score $S_L(x)$ to obtain the final score $S(x)$. Specifically, the definition of $S_L(x)$ is as follows:
%Afterwards, we can regularize the score we obtained from the object detector 
% $S_D(x)$ 
%for an object hypothesis $x$ as follows:\par
\begin{equation}
S_L(x)=\left\{
\begin{array}{cl}
-1  &\quad    M(x) < 0.15 \\
 0  &\quad    0.15 < M(x) < 0.6 \\
\alpha e^{M(x)} + b &\quad    M(x) \geq 0.6 \\
\end{array} \right.
\label{eqn1}
\end{equation}
Here $S_L (x)$ is the scene layout score of object $x$. $M(x)$ is the score based on the location of object $x$ in the scene layout that includes both obstacle and road distributions, 
%Here $S_L (x)$ is the detection score of object $x$ after regularization. $M(x)$ is the scene layout score that includes both obstacle and road distributions, 
% $S_D (x)$ is the object detection score produced by the detector, 
${\alpha}$ is a variable parameter to adjust the score of the scene layout, and $b$ is a fixed bias value of the scene layout score.\par
According to Eqn.~\ref{eqn1}, we assume that when the score $M(x) < 0.15$, the object is outside the road distribution, which is considered a false positive. When the score is $0.15 < M(x) < 0.6$, we consider that the object is within the road region but close to the boundary, so there will be no change to the detection score. When the score ${M(x) \geq 0.6}$, it is considered that the object appears at a position that we are most likely to detect the obstacles, and the score is increased accordingly.\par

\subsection{Obstacle Tracking with Optical Flow} 
Here, we use the Lucas-Kanade (LK) method~\cite{bouguet2001pyramidal} to calculate the optical flow between two consecutive frames, and note that other methods may also be used. Specifically, we consider two frames from time $t$ to ${t + \delta}$, in order to achieve the tracking of the detected objects. In general, LK method is based on the following three assumptions: (1) the brightness of the tracked part of the object in the scene remains basically unchanged; (2) the motion is relatively slow relative to the frame rate; (3) adjacent points on the same surface in a scene should have similar motion. In addition, the image pyramid is introduced to improve its performance. By reducing the size of the image, the moving speed of the object in the image is relatively reduced, so that the object with faster moving speed can be tracked with better quality.\par
\begin{figure}
\includegraphics[width=\textwidth]{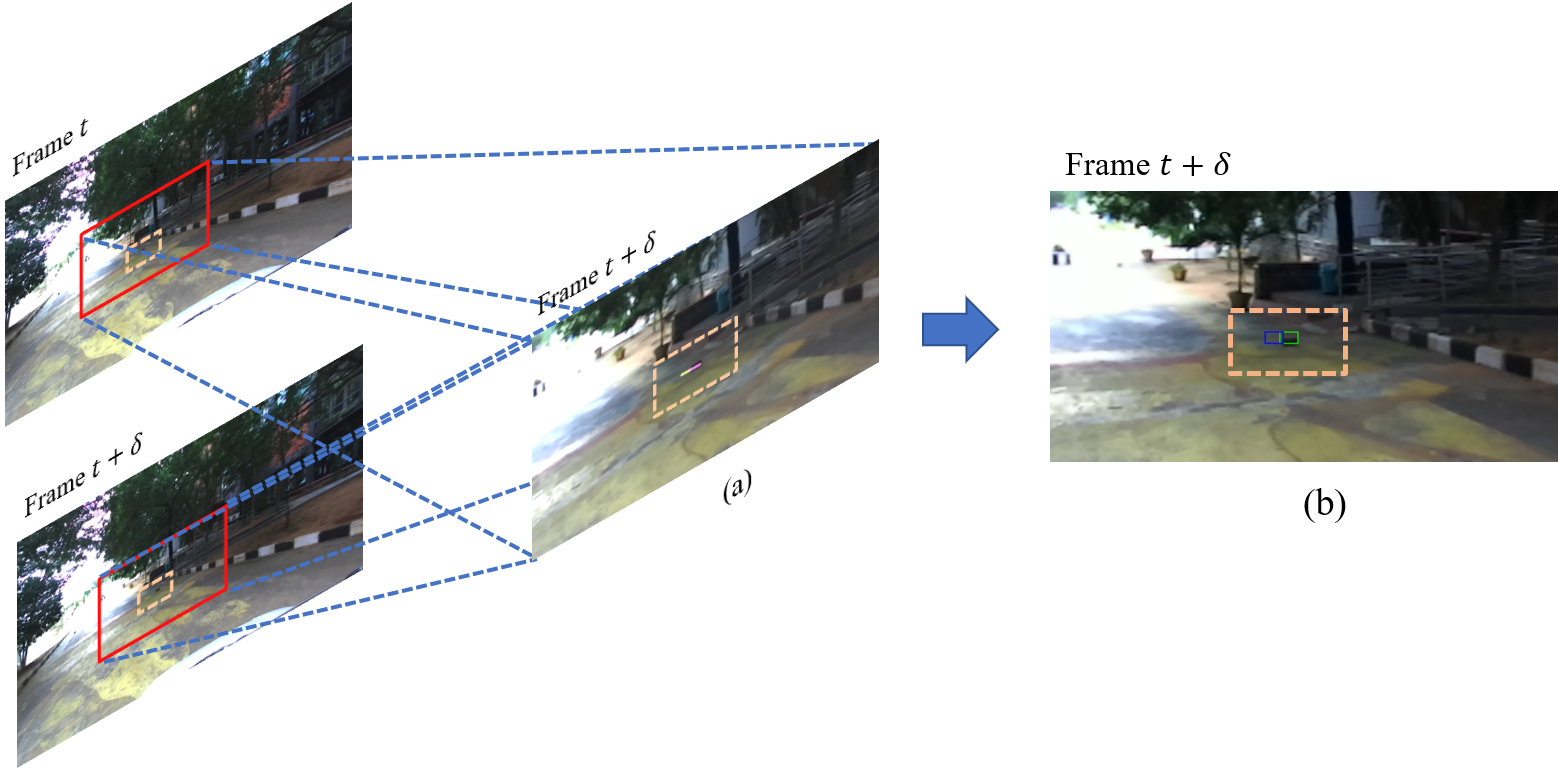}
\caption{Obstacle tracking with optical flow. The orange dashed box indicates the object area used to calculate the optical flow of adjacent frames. (a) The yellow and pink lines  represent the offset of the object. (b) The blue box indicates the bounding box overlaid from the preceding frame. The green box indicates the bounding box obtained by optical flow tracking in the following frame. Best viewed in electronically, zoomed in.} \label{fig4}
\end{figure}

%the relationship between frames obtained by vehicle monitoring during vehicle driving can better meet the application conditions of optical flow algorithm. Therefore, we can well calculate the offset of obstacles relative to vehicles, so as to realize achieve tracking, as illustrated in Fig.~\ref{fig4}. The specific steps are as follows:\par
While detecting and tracking road obstacles, the assumptions above may not be met due to constantly changing motion, illumination, partial occlusion, etc. Therefore, instead of calculating the optical flow for the entire scene, we focus on the vicinity where an object is detected. As such, we can eliminate the unnecessary distractions from other parts of the scene, as illustrated in Fig.~\ref{fig4}. The specific steps are as follows:\par
\paragraph{Step 1.} The detector is used to detect the obstacles, obtain the detection results of the preceding and the following frames, and select the detections with scores $ S_D(x) > 0.3$ in the preceding frame.
\paragraph{Step 2.} To judge whether the detection is missed in the following frame, we define a search area $A_r$ in the following frame by enlarging the detection bounding box of the detected object. The definition of the search area $A_r$ is given as: 
\begin{equation}
    A_r(x) = (b_x, b_y, \alpha b_w+\tau, \beta b_h+ \tau)
\label{eqn2}
\end{equation}
\noindent where $\alpha$ and $\beta$ are the adjustment coefficients of the selected area width and height respectively. $\tau$ represents the initial size of the area. Here, $\alpha > 1$, $\beta > 1$, and $\tau > 0$ as we only enlarge the bounding box to define the search area.
\paragraph{Step 3.} If the obstacle is missed in the following frame (i.e., no bounding box for the same obstacle category within $A_r(x)$), we crop out the the search area given in step 2 in both the preceding and the following frames, and obtain image corners in the cropped area using the Shi-Tomasi corner detection algorithm~\cite{tommasini1998making}, and select the feature corners located on the detected object.
\paragraph{Step 4.} The Lucas Kanade optical flow method is then used to track the feature corners in the cropped area, in order to obtain the offset of the feature corners in the two frames before and after.
\paragraph{Step 5.} We transfer the bounding boxes in the previous frame to the following frame through the obtained offset. Again, this happens only when the obstacle is missed in the following frame.

\subsection{Spatio-Temporal Aware Detection} 
Based on the spatial and temporal context modeling above, we combine them in the final inference process. Firstly, the obstacles in the input image are detected by the detector, and then the scene layout model is used to process the detection results to remove the incorrect detections outside the road scene distribution, and strengthen the confidence of small or weak obstacles in the road region (see Sec. 3.1). Secondly, the bounding box of missed detections in the road region is recovered by obstacle tracking and bounding box transfer (see Sec. 3.2). Finally, in order to avoid false positives, we calculate the score of the recovered bounding boxes in combination with the scene layout model. Our final score function is written as follows:
\begin{equation}
{ S(x_{(t+\delta)}) = S(x) - (\lambda \log M^2(x_{(t+\delta)}) + b) }
\label{finalSorceByGen}
\end{equation}
\noindent where $S(x_{(t+\delta)})$ is the score function for object $x_{(t+\delta)}$ generated by optical flow tracking. 
%$M(x_{(t+\delta)})$ is the scene layout score of $x_{(t+\delta)}$ (see Sec. 3.1).
$M(x_{(t+\delta)})$ is the location score of $x_{(t+\delta)}$ in the scene layout (see Sec. 3.1).
${\lambda}$ is a variable parameter used to adjust the score of the scene layout model. ${b}$ is a bias, giving a fixed initial score to the scene layout model. If the score $ S(x_{(t+\delta)}) < 0.3$, we treat the recovered bounding box as invalid and remove it.

\section{Experiments}
\subsection{Datasets} 
Our experiments are carried out on two public datasets: the Small Obstacle Detection (SOD) dataset~\cite{singh2020lidar} and the Lost and Found dataset~\cite{pinggera2016lost}, for evaluating the efficacy of the method. There are 2927 images in the SOD dataset, including 1937, 530 and 460 images in the training, validation, and test set, respectively. It comprises of 15 video sequences in total and utilizes a diverse set of small obstacle instances, and uses different road scenes and different sets of obstacles while recording the train, val and test sequences. Test split is kept to be most challenging in terms of turns, occlusions and shadows to better evaluate the generalization ability. The Lost and Found dataset contains images of small items, such as cargo, wooden strips and toys scattered in the free space in front of the car. Among them, training set and test set contain 1036 and 1068 images.\par

\subsection{Implementation Details} 
In this work, we use Faster RCNN~\cite{ren2015faster} and YOLOv5~\cite{jocher2021ultralytics} object detection algorithms as our baselines. We use their latest implementations in PyTorch without any changes. In the Faster RCNN algorithm, we use ResNet-50~\cite{he2016deep} as the backbone network, SGD as the optimizer, and use a minibatch size of $8$. A total of $30$ epochs were trained. The learning rate starts from $0.006$, and the learning rate decreases to one-third of the original every $5$ epochs. We also use the ImageNet~\cite{deng2009imagenet} pre-trained model for network initialization. In addition, we use feature pyramid networks (FPN)~\cite{lin2017feature} in order to learn high quality multi-scale feature representations. In the YOLOv5 algorithm, we use the yolov5l6 network structure, and use the MS-COCO~\cite{lin2014microsoft} pre-trained model to initialize the network weight. SGD is used as the optimizer, and the minibatch size is set to $8$. A total of $100$ epochs are trained, and the epoch with the best validation results is selected.
In addition, for the search area in Eqn.~\ref{eqn2}, we set $~\alpha = 2,~\beta = 3$, $\tau = 30$ and $\tau = 40$ for SOD and Lost and Found, respectively.\par
%The other parameters use the default configuration.
%At post-processing stage, the input size of images are the same as the one in training. The predictions with classification scores $s > 0.05$ are selected for evaluation.\par

\subsection{Experimental Results}
\subsubsection{Metrics} 
Following common practice~\cite{everingham2010pascal,lin2014microsoft}, we use average precision (AP) to evaluate the detection algorithm. Specifically, true positives (TP) represent the number of positive objects correctly detected, false positives (FP) represent the number of background regions incorrectly marked as objects, false negatives (FN) represent the number of positive objects not detected, and recall and precision are calculated according to Eqns.~\ref{recall} and~\ref{precision}.
\begin{equation}
Recall = \frac{TP}{TP + FN}
\label{recall}
\end{equation}
\begin{equation}
Precision = \frac{TP}{TP + FP}
\label{precision}
\end{equation}\par
%curve with recall as the abscissa and precision as the ordinate to obtain
Given recall and precision at varying detection score thresholds, we could plot the precision-recall curve of the test results, and the average accuracy (AP) represents the area between the curve and the coordinate axes. In addition, we evaluate results computed at varying Intersection over Union (IoU)~\cite{everingham2010pascal} thresholds, i.e., AP values at 50$\%$, 75$\%$ IoU and the average of AP from 50$\%$ to 95$\%$ IoU (at 5$\%$ IoU interval) as the evaluation metrics.\par
\subsubsection{Results} 
Taking Faster RCNN and YOLOv5 as our baselines, the experimental results obtained by adding the scene layout model (SL), the obstacle tracking with optical flow (OF) and the combination of the two methods are shown in Table~\ref{tab1}.\par

\begin{table}[!htbp] 
\centering
\caption{The ablation study on SOD dataset. $\textbf{SL:}$  Scene layout. $\textbf{OF:}$  Optical flow tracking. Note that our method improves performance both the Faster RCNN and the YOLOv5 baselines.}\label{tab1}
\setlength{\tabcolsep}{2.3mm}{
\begin{tabular}{lllllll} %需要10列
\hline
Method  & AP$_{50}$  &AP$_{75}$ &AP &AP$_{S}$ &AP$_{M}$ &AP$_{L}$\\
\hline
Faster RCNN &48.4&21.8&24.6&22.8&57.3&23.3\\
Faster RCNN+SL &50.9&$\mathbf{22.2}$&25.3&23.1&59.9&23.3\\
Faster RCNN+OF &49.8&22.0&25.0&23.0&57.5&23.3\\
Faster RCNN+SL+OF &$\mathbf{52.6}$&22.1&$\mathbf{25.7}$&$\mathbf{23.4}$&$\mathbf{60.1}$&$\mathbf{23.3}$\\
\hline
YOLOv5 &57.8&27.7&30.7&29.3&57.7&27.3\\
YOLOv5+SL &58.3&27.8&30.9&29.2&57.9&27.1\\
YOLOv5+OF &58.9&$\mathbf{27.9}$&31.1&29.4&$\mathbf{58.2}$&$\mathbf{28.9}$\\
YOLOv5+SL+OF &$\mathbf{59.4}$&27.7&$\mathbf{31.2}$&$\mathbf{29.6}$&58.1&27.9\\
\hline
\end{tabular}}
\end{table}

\begin{figure}
\includegraphics[width=\textwidth]{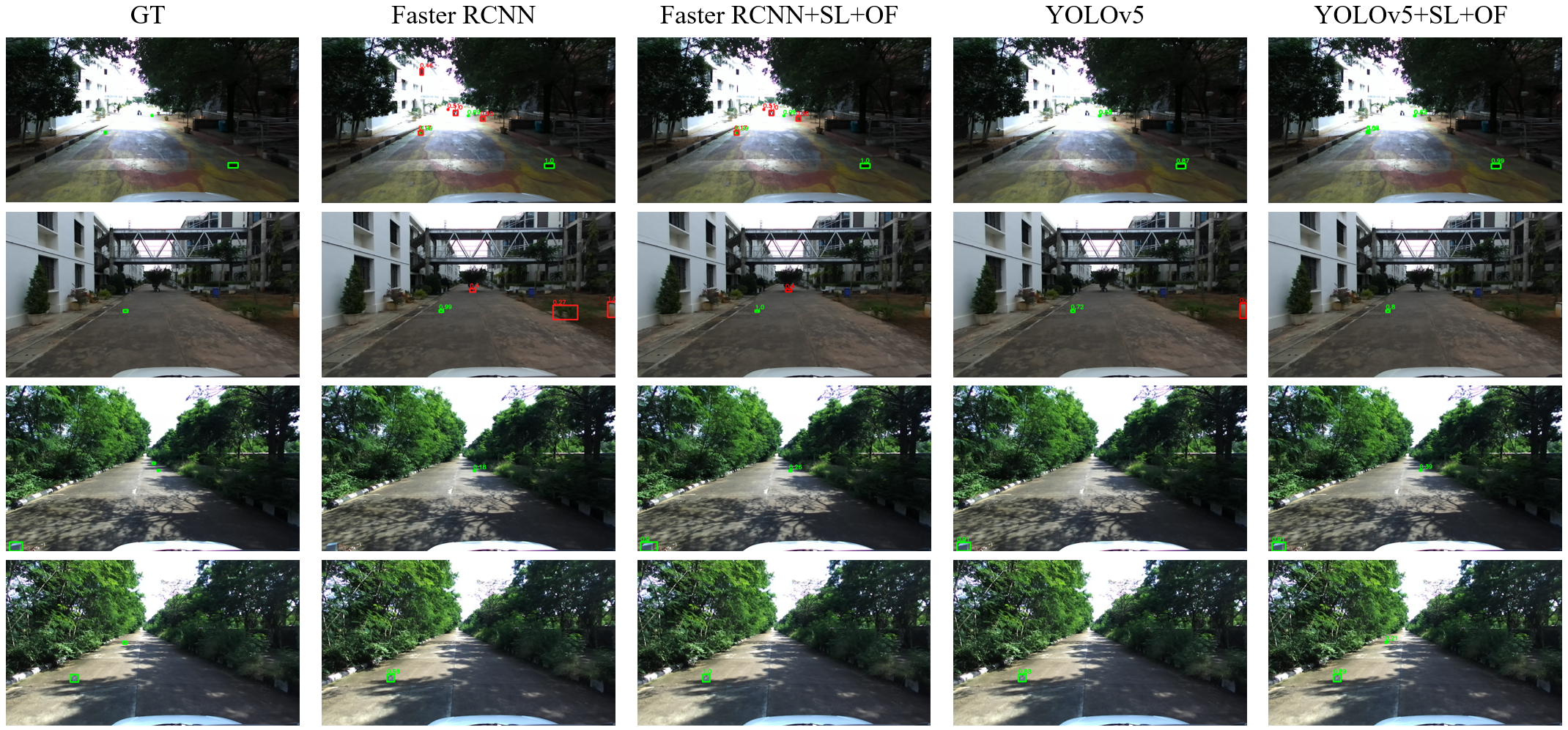}
\caption{Example detection results on the test set of the SOD dataset. Columns: $\textbf{GT:}$ Input image with ground-truths overlaid. $\textbf{Faster RCNN:}$ Detections with Faster RCNN. $\textbf{Faster RCNN+SL+OF:}$ Detections with Faster RCNN+Scene Layout+Optical Flow. $\textbf{YOLOv5:}$ Detections with YOLOv5. $\textbf{YOLOv5+SL+OF:}$ Detections with YOLOv5+Scene Layout+Optical Flow. Red boxes are false positives, green boxes are true positives. Detection score threshold is $0.05$. Best viewed electronically, zoomed in.} \label{fig5}
\end{figure}

Taking AP$_{50}$ as an example, the average precision is increased by 2.5$\%$ and 1.4$\%$ when the scene layout (SL) model and object tracking with optical flow (OF) method are used separately with Faster RCNN, which demonstrates the efficacy of the two methods. Combining these two methods provides a significant improvement by 4.2$\%$. For YOLOv5, the scene layout model provides a modest improvement of 0.5$\%$. However, object tracking with optical flow provides a considerable improvement of 1.1$\%$. Combining these two methods provides the largest improvement of 1.6$\%$. We also present some qualitative results in Fig.~\ref{fig5}, which clearly show that our method is able to eliminate false positives while being able to detect distant obstacles with weak visual support.\par

\begin{figure}
\includegraphics[width=\textwidth]{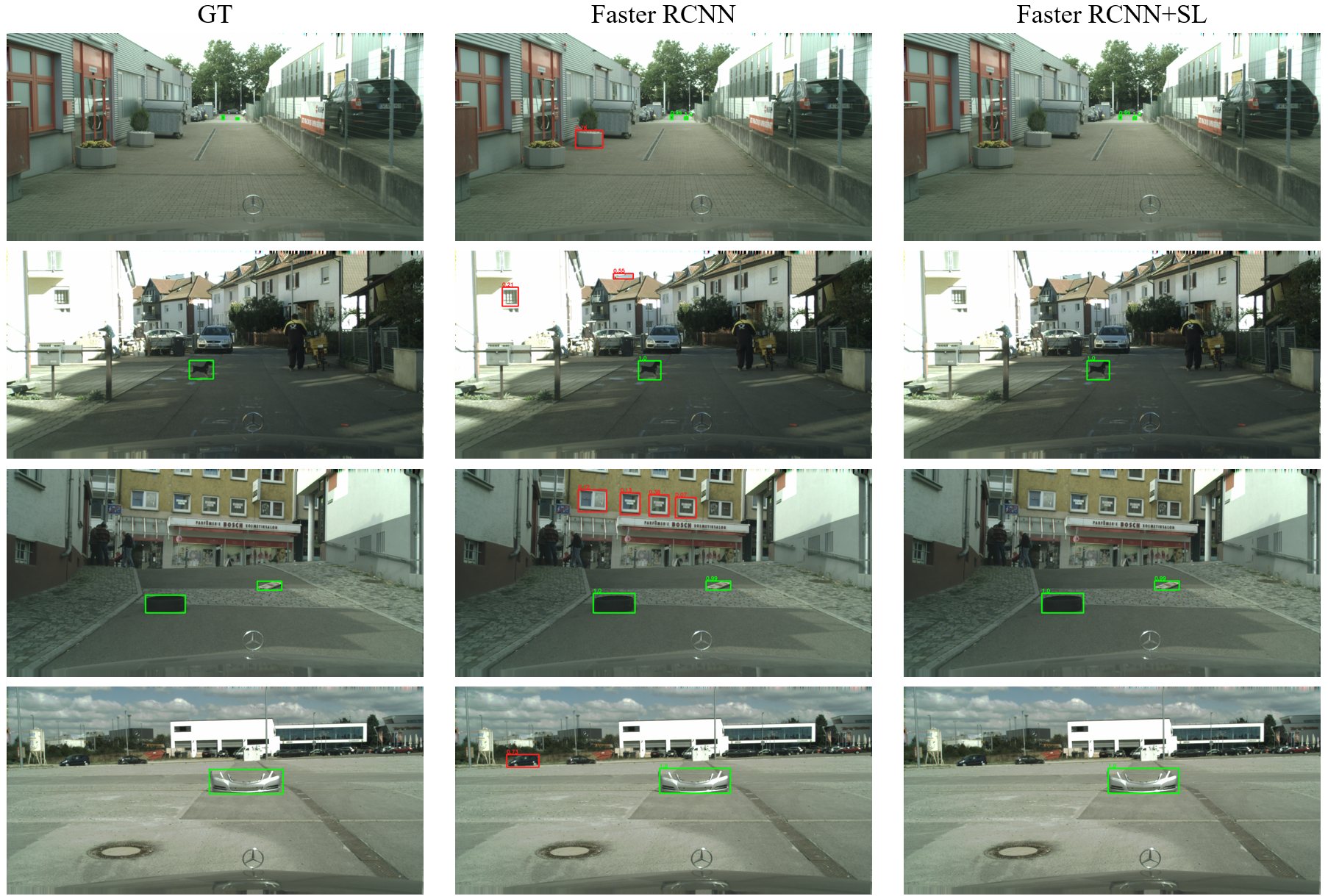}
\caption{Example detection results on the test set of the Lost and Found dataset. Columns: $\textbf{GT:}$ Input image with ground-truths overlaid. $\textbf{Faster RCNN:}$ Detections with Faster RCNN. $\textbf{Faster RCNN+SL:}$ Detections with Faster RCNN+Scene Layout. Red boxes are false positives, green boxes are true positives. Detection score threshold is $0.05$. Best viewed electronically, zoomed in.} \label{fig6}
\end{figure}

In addition, due to the much larger time interval between adjacent image frames in the Lost and Found dataset, the temporal smoothness between two adjacent frames is poor. Therefore, in this dataset, we only use the scene layout model. The quantitative results we obtained are summarized in Table ~\ref{tab2}. It can be seen that $ AP_{50} $ is increased by 0.9$\%$ when the scene layout model is used. We also show some qualitative results in Fig.~\ref{fig6}, demonstrating the ability of our method to remove false positives, while boosting the scores of obstacles in the scene layout detected by Faster RCNN.\par

\begin{table}[!htbp] 
\centering
\caption{Results with and without scene layout model on Lost and Found dataset.}\label{tab2}
\setlength{\tabcolsep}{2.3mm}{
\begin{tabular}{lllllll} %需要10列
\hline
Method  & AP$_{50}$  &AP$_{75}$ &AP &AP$_{S}$ &AP$_{M}$ &AP$_{L}$\\
\hline
Faster RCNN &65.4&39.9&38.3&28.7&52.0&67.0\\
Faster RCNN+SL &$\mathbf{66.3}$&$\mathbf{40.4}$&$\mathbf{38.7}$&$\mathbf{28.8}$&$\mathbf{52.7}$&$\mathbf{67.9}$\\
\hline
\end{tabular}}
\end{table}


The computational efficiency of our method is also acceptable for practical applications. Taking the baseline Faster RCNN algorithm on SOD dataset as an example, the average inference time per image is about 61ms on a single nVIDIA RTX 3080 Ti. The inference time does not change when we introduce the spatial context, as the parameters in the scene layout model can be calculated in advance and used directly in the inference phase. Then, with the optical flow tracking added, the processing time increases to 89ms per image. It should be noted that because we adopt the optical flow calculation method based on region selection to eliminate unnecessary computations in a large part of the image, the additional computational budget for the temporal context is reduced to only 28ms per image.\par

Based on the analysis above, both the spatial context and the temporal context improve final detection performance when used alone, and their combination produces the best detection results. These results validate the efficacy of the proposed method.

% \begin{figure}
% \includegraphics[width=\textwidth]{r_test_2.png}
% \caption{Example detection results on our held-out test set of the Lost And Found dataset. Columns: $\textbf{GT:}$ Ground-truth. $\textbf{Faster RCNN:}$ Detections with Faster RCNN. $\textbf{Faster RCNN+heatmap:}$ Detections with Faster RCNN+heatmap. Red box is the false positive, green box is true positive.} \label{fig6}
% \end{figure}

\section{Conclusion}
In this paper, we propose a novel method to detect road obstacles by integrating the spatial and temporal context. Specifically, we derive the spatial distribution of obstacles and the road with a data-driven approach to build a scene layout model, and propose an obstacle tracking method based on optical flow and object region selection to encode temporal smoothness while suppressing the adverse impact of background noise. Our experiments show significant improvements in object detection accuracy compared to state-of-the-art baseline detectors. As a general framework for spatio-temporal context modeling, our method can work with other object detection algorithms not mentioned in this paper. In the future, we plan to further integrate spatio-temporal context modeling into deep models to allow for end-to-end training of a unified model.

\vspace{5mm}

\footnotesize{\noindent \textbf{Acknowledgments.} 
This work is partially supported by NSFC (61972187, 61703195), Fujian NSF (2022J011112, 2020J02024), the Open Program of The Key Laboratory of Cognitive Computing and Intelligent Information Processing of Fujian Education Institutions, Wuyi University (KLCCIIP2020202), and the Research Startup Fund of Minjiang University (MJY19021). Lingyu Liang is supported by Science and Technology Program of Guangzhou (202102020692), the Open Fund of Ministry of Education Key Laboratory of Computer Network and Information Integration (Southeast University) (K93-9-2021-01), the Open Fund of Fujian Provincial Key Laboratory of Information Processing and Intelligent Control (Minjiang University) (MJUKF-IPIC202102), Guangdong NSF (2019A1515011045) and CAAI-Huawei MindSpore Open Fund.}

{\small
\bibliographystyle{ieee}
\bibliography{ref}
}

\end{document}